\documentclass{article}
\usepackage{ijcai23}

\usepackage{amsmath,amsfonts,bm}

\def\eqref#1{equation~\ref{#1}}

\def\1{\bm{1}}

\def\ve{{\bm{e}}}

\def\vh{{\bm{h}}}

\def\vm{{\bm{m}}}

\def\vu{{\bm{u}}}
\def\vv{{\bm{v}}}

\def\vx{{\bm{x}}}

\def\mX{{\bm{X}}}

\DeclareMathAlphabet{\mathsfit}{\encodingdefault}{\sfdefault}{m}{sl}
\SetMathAlphabet{\mathsfit}{bold}{\encodingdefault}{\sfdefault}{bx}{n}

\def\gB{{\mathcal{B}}}

\def\gE{{\mathcal{E}}}

\def\gG{{\mathcal{G}}}

\def\gV{{\mathcal{V}}}

\def\sG{{\mathbb{G}}}

\def\sL{{\mathbb{L}}}

\def\sR{{\mathbb{R}}}

\usepackage{times}
\usepackage{soul}
\usepackage{url}
\usepackage[hidelinks]{hyperref}
\usepackage[utf8]{inputenc}
\usepackage[small]{caption}
\usepackage{graphicx}
\usepackage{amsmath}
\usepackage{amsthm}
\usepackage{booktabs}
\usepackage{algorithm}
\usepackage[switch]{lineno}
\usepackage{pifont}
\newcommand{\whiteding}[1]{\ding{\numexpr171+#1\relax}}

\usepackage{booktabs}       
\usepackage{amsfonts}       
\usepackage{nicefrac}       
\usepackage{microtype}      
\usepackage{xcolor}         
\usepackage{microtype}
\usepackage{graphicx}
\usepackage{subfigure}
\usepackage{enumitem}
\usepackage{mhchem}
\usepackage{color}
\usepackage{mathtools}
\usepackage{wrapfig, blindtext}
\usepackage{subfigure}
\usepackage{multirow}
\usepackage{makecell}
\usepackage{array, booktabs, xcolor}
\usepackage{amssymb}
\usepackage{graphbox}
\usepackage{algpseudocode}
\usepackage{caption}
\allowdisplaybreaks

\newcommand{\ie}{\textit{i}.\textit{e}.}
\newcommand{\eg}{\textit{e}.\textit{g}.}

\title{MolKD: Distilling Cross-Modal Knowledge in Chemical Reactions\\for Molecular Property Prediction}

\author{
Liang Zeng\textsuperscript{\rm 1},
Lanqing Li\textsuperscript{\rm 2,3}\thanks{Corresponding authors.},
Jian Li\textsuperscript{\rm 1}\footnotemark[1]
\affiliations
\textsuperscript{\rm 1}Institute for Interdisciplinary Information Sciences (IIIS), Tsinghua University\\
\textsuperscript{\rm 2}Zhejiang Lab, \textsuperscript{\rm 3}The Chinese University of Hong Kong\\
zengl18@mails.tsinghua.edu.cn, lanqingli1993@gmail.com, lijian83@mail.tsinghua.edu.cn
}

\begin{document}

\maketitle

\begin{abstract}
How to effectively represent molecules is a long-standing challenge for molecular property prediction and drug discovery. 
This paper studies this problem and proposes to incorporate chemical domain knowledge, specifically related to chemical reactions, for learning effective molecular representations. 
However, the inherent cross-modality property between chemical reactions and molecules presents a significant challenge to address.
To this end, we introduce a novel method, namely MolKD, which \underline{D}istills cross-modal \underline{K}nowledge in chemical reactions to assist \underline{Mol}ecular property prediction. 
Specifically, the reaction-to-molecule distillation model within MolKD transfers cross-modal knowledge from a pre-trained teacher network learning with one modality~(\ie, reactions) into a student network learning with another modality~(\ie, molecules).
Moreover, MolKD learns effective molecular representations by incorporating \emph{reaction yields} to measure transformation efficiency of the reactant-product pair when pre-training on reactions. Extensive experiments demonstrate that MolKD significantly outperforms various competitive baseline models, \eg, $2.1\%$ absolute AUC-ROC gain on Tox21. Further investigations demonstrate that pre-trained molecular representations in MolKD can distinguish chemically reasonable molecular similarities, which enables molecular property prediction with high robustness and interpretability.
\end{abstract}
\section{Introduction}
\label{sec:introduction}
Effective molecular representations are crucial prerequisites in AI-aided drug design and discovery, such as chemical reaction prediction~\cite{lu2022unified,wang2021chemical}, molecular property prediction~\cite{stark20223d,liu2022structured,rong2020self}, and molecule generation~\cite{xu2022geodiff,luo2021autoregressive}.
Recently, many works have exploited graph neural networks~(GNNs) to obtain effective molecular representations by regarding the topology of atoms and bonds as a graph, and propagating messages of each atom within its neighbors. 
With the preservation of rich structural information in molecules, these methods have shown promising results in molecular property prediction~\cite{stark20223d,fang2022geometry,li2022imdrug}.
However, acquiring labeled molecule data usually requires time-consuming laborious and costly wet-lab experiments~\cite{atz2021geometric,ghiandoni2019development}, which 
hinders the successful application of GNN-based methods in molecular property prediction.

To mitigate the scarcity of labeled molecule data, recent progress has been made by pre-training molecular representations on enormous unlabelled molecule data.
\cite{wang2022molecular,liu2022structured} design several dedicated self-supervised learning tasks to model molecular structures. \cite{fang2022geometry,stark20223d} introduce 3D molecular information and align 2D molecular graphs with their 3D molecular conformers to obtain effective representations. 
However, these methods may suffer from low data efficiency and generalization ability without the help of chemical domain knowledge~\cite{munkhdalai2015incorporating,vagin2004refmac5}. To this end, we propose to incorporate chemical domain knowledge, specifically related to chemical reactions, for learning effective molecular representations.

So far, only a few works have been proposed to effectively leverage chemical domain knowledge for molecular representation learning. This is a long-standing challenge for molecular property prediction and drug discovery. KCL~\cite{fang2022molecular} constructs a chemical element knowledge graph to incorporate chemical domain knowledge into graph semantics and model the correlations between atoms that have common attributes. MolR~\cite{wang2021chemical} preserves the equivalence of molecules between reactants and products with respect to chemical reactions in the embedding space. 
However, existing works have mainly overlooked the inherent modality gap between chemical reactions and molecules.
In this paper, we push these previous works one step further by proposing a novel method~(MolKD), which performs cross-modal \underline{K}nowledge \underline{D}istillation from chemical reactions into molecules to assist \underline{Mol}ecular property prediction. 

To narrow the gap in data modality between reactions and molecules, we introduce the reaction-to-molecule distillation model to transfer cross-modal knowledge from a pre-trained teacher network learning with one modality~(\ie, reactions) into a student network learning with another modality~(\ie, molecules). We adopt the feature-based contrastive representation distillation method to facilitate knowledge transfer between reactions and the downstream molecular property prediction tasks.
Moreover, for learning powerful molecular representations, we improve upon the chemical-reaction-aware pre-training method~\cite{wang2021chemical} by incorporating \emph{reaction yields}, which play a critical role in chemical synthesis by measuring transformation efficiency of the reactant-product pair.
We propose to adaptively scale the margin of the pre-training loss in terms of reaction yields.

To validate the high quality of molecular representations, we compare our yield-guided chemical-reaction-aware pre-training method with a set of competitive baselines, and achieve $2.8\%$ absolute Hit@1 gain on USPTO in chemical reaction prediction. 
We also visualize the selected molecules to show that the learned molecular representations are chemically meaningful by encoding structural and synthetic semantics in the representation domain~(See Fig.~\ref{fig:molecule-query}).
To verify the effectiveness of our proposed MolKD, we compare it with several competitive baselines on 10 molecular property prediction benchmarks, among which MolKD achieves superior performance on 8 challenging tasks, \eg, $2.1\%$ absolute AUC-ROC gain on Tox21. 
We further investigate MolKD on PhysProp~\cite{li2022glam} to demonstrate its robustness and interpretability.
In summary, our main \textbf{contributions} are:
\begin{itemize}[leftmargin=10pt]
    \item We present MolKD, to our knowledge, the first novel method designed for distilling cross-modal pre-trained chemical knowledge in reactions to assist molecular property prediction.
    \item We propose to incorporate reaction yields as a measure of transformation efficiency and demonstrate its effectiveness in assisting pre-training on reactions.
    \item We evaluate the effectiveness of MolKD thoroughly on various molecular property prediction datasets. Experimental results demonstrate that MolKD significantly outperforms competitive baselines on multiple benchmarks.
\end{itemize}
\section{Related work}
\label{sec:related}
\paragraph{Knowledge Distillation.} Knowledge distillation is a generic training paradigm where a student model is trained under the supervision of a teacher model to achieve~(implicit) knowledge transfer~\cite{gou2021knowledge}. According to~\cite{tian2019contrastive}, there mainly exist three distillation paradigms: (1) model compression~\cite{hinton2015distilling}, (2) cross-modal knowledge transfer~\cite{romero2014fitnets}, (3) ensemble distillation~\cite{bucilua2006model}.
We can also classify distillation techniques from the perspective of knowledge categories~\cite{gou2021knowledge}: logit-based~\cite{hinton2015distilling} and feature-based~\cite{tian2019contrastive} distillation method. 
In this paper, we focus on the cross-modal (from chemical reaction data into molecule data) and feature-based~(since it was reported to achieve superior performance than logits-based methods) distillation method. 
To our best knowledge, this is the first work to explore the reaction-to-molecule cross-modal knowledge distillation method for molecular property prediction.

\paragraph{Pre-training for Molecular Representations.} In general, the pre-training methods for molecular representations fall into three categories.
(1) \cite{wang2022molecular,liu2022structured,fang2022molecular,hu2019strategies,rong2020self,li2020learn} design dedicated self-supervised learning tasks based on string-level~(1D) and graph-level~(2D) structures of molecules.
\cite{hu2019strategies} propose the node-~(attribute masking) and graph-level~(context prediction) pre-training strategies to make accurate and robust predictions on downstream tasks. 
GROVER~\cite{rong2020self} adopts the Transformer structure to the designed self-supervised tasks~(\ie, contextual property prediction and graph-level motif prediction).
(2) \cite{fang2022geometry,stark20223d,liu2021pre} further introduce 3D conformations of a molecule and align its 2D and 3D representation.
GraphMVP~\cite{liu2021pre} introduces a multi-view pre-training framework to preserve consistency between 2D topological structures and 3D geometric views.
(3) \cite{wang2021chemical,fang2022molecular} propose to improve molecular representations using extra data sources. MolR~\cite{wang2021chemical} adopts the composable TransE methods in NLP to preserve the equivalence of molecules with respect to chemical reactions in the embedding space. KCL~\cite{fang2022molecular} constructs a chemical element knowledge graph to incorporate prior knowledge into molecular graph semantics.
Different from MolR, we incorporate the important factor of chemical reactions---yields---as a measure of transformation efficiency to adaptively scale the margin of the TransE loss in the pre-training phase and obtain more informative molecular representations.

\section{Background}
\label{sec:background}
\paragraph{Molecular Graphs.} Molecules can be naturally represented as graphs by taking atoms as nodes and chemical bonds as edges. Formally, let $\gG=\left(\gV, \gE, \mX\right)$ denote a molecular graph, where $\gV = \left\{v_i\right\}_{i=1}^N$ is a set of $N$ atoms, and $\gE \subseteq \gV \times \gV$ is a set of chemical bonds between atoms. $\mX = \left[\vx_1, \vx_2, \cdots, \vx_N\right]^{T}$ $\in \sR^{N \times d}$ represents the atom feature matrix, where $d$ is the feature dimension. Each atom $i$ has the an initial feature vector $\vx_i \in \sR^d$, such as molecule fingerprints~\cite{wu2018moleculenet}. 

\begin{figure*}[h]
    \centering
    \includegraphics[width=1.\linewidth]{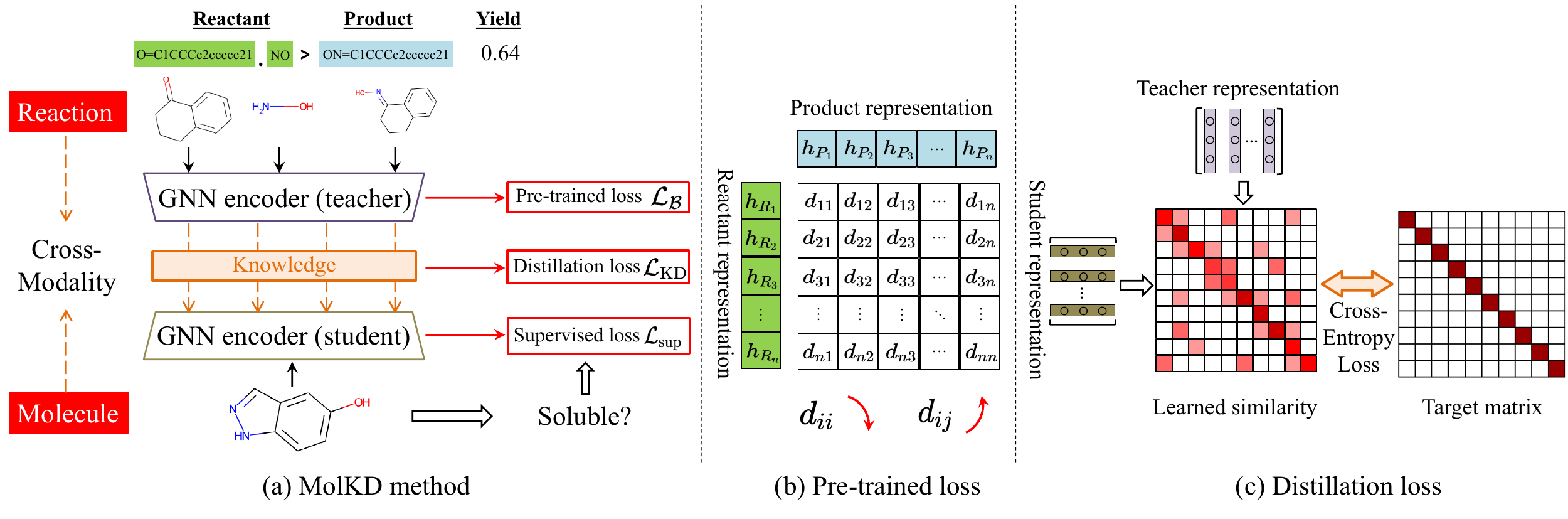}
    \caption{An overview of our MolKD method. MolKD consists of two-stage training models: the reaction-to-molecule knowledge distillation model and the yield-guided chemical-reaction-aware pre-training model.}
    \vspace{-1em}
    \label{fig:framework}
\end{figure*}

\paragraph{Chemical Reactions.} In this paper, a chemical reaction can be described as a pair of molecular graphs associated with a reaction yield $(\gG_R, \gG_P, y)$, where $\gG_R = \{ \gG_{r_1}, \gG_{r_2}, \cdots\}$ and $\gG_P = \{ \gG_{p_1}, \gG_{p_2}, \cdots\}$ are the set of reactant graphs and product graphs respectively, and $y$ is the yield of this reaction.
In organic chemical synthesis processes, the reaction yield is a primary factor, which measures the amount of a specific set of products formed per mole of the reactants consumed~\cite{fogler1999elements}. Yields are usually expressed as a percentage~(ratio of actual yield to theoretical yield) and fall into the range $[0, 1]$. According to~\cite{fogler1999elements}, yields above $70\%$ are relatively good due to side and incomplete reactions that generate other products. In general, the higher yield indicates that the reaction is more efficient and important for organic chemical synthesis in practice.

\paragraph{Uncertainty Knowledge Graphs~(KG).} An uncertainty KG consists of a set of weighted triples $\{(l, s_l)\}$~\cite{chen2021probabilistic}. For each pair $(l, s_l)$, $l=(h,r,t)$ is a triple representing a relation fact where $h$ is the head entity, $t$ is the tail entity, and $r$ is the associated relation. $s_l \in [0,1]$ denotes the confidence score for this relation fact to be true. Some examples of weighted triples are as follows~\cite{kertkeidkachorn2019gtranse}: ((university, \emph{synonym}, institute), 0.86) and ((fork, \emph{atlocation}, kitchen), 0.4). Overall, given the uncertainty KG, we aim to learn an embedding model to encode each entity and relation into a low-dimensional space where confidence scores of relation facts are preserved. In Sec.~\ref{sec:pre-train}, we utilize the confidence score of uncertainty KG to model transformation efficiency of chemical reactions with yields.

\paragraph{Graph Neural Networks~(GNNs).} GNNs have become increasingly popular in various molecular modeling tasks~\cite{wu2020comprehensive}. Typically, the training process of GNNs follows the message-passing scheme~\cite{hamilton2020graph}. During each message-passing iteration $k$, a hidden embedding $\vh_u^{(k)}$ corresponding to each node $u \in \gV$ can be expressed as follows:
\begin{small}
\begin{equation}
    \begin{aligned}
    \vh_{u}^{(k)} &=\text { UPDATE }^{(k)}\left(\vh_{u}^{(k-1)}, \text { AGG }^{(k)}\left(\left\{\vh_{v}^{(k-1)}, v \in \mathcal{N}(u)\right\}\right)\right) \\
    &=\text { UPDATE }^{(k)}\left(\vh_{u}^{(k-1)}, \vm_{\mathcal{N}(u)}^{(k)}\right),
    \end{aligned}
\end{equation}
\end{small}
where $\text{UPDATE}$ and $\text{AGG}$ are arbitrary differentiable functions~(\ie, neural networks), and $\vm_{\mathcal{N}(u)}$ denotes the ``message'' that is aggregated from $u$'s neighborhood $\mathcal{N}(u)$. The initial embedding at $k=0$ is set to the input features,~\ie, $\vh_u^{(0)} = x_u, \forall u \in \mathcal{V}$. After running $k$ message-passing iterations, we can obtain information from $k$-hops nodes. 
Different GNNs can be obtained by choosing different $\text{UPDATE}$ and $\text{AGG}$ functions. 
For graph classification tasks, a graph-level representation $\vh_{\gG}$ is obtained by integrating all the node embeddings $\vh_{u}^{(K)}$ among the graph $\gG$ after $K$ iterations:
\begin{equation}
    \vh_{\gG} = \text{READOUT} \left(\left\{ \vh_{u}^{(K)}, u \in \mathcal{V} \right\}\right),
\end{equation}
where the READOUT$(\cdot)$ is a permutation invariant function such as summation and maximization operators or more complex pooling methods~\cite{hu2019strategies}.

\paragraph{Problem Setup.} For molecular property prediction, given a training set of molecular graphs $\{ \gG_1, \gG_2, \cdots, \gG_{N_l}\} \subseteq \sG$ and their labels $\{l_1, l_2, \cdots, l_{N_l}\} \subseteq \sL$, \eg, solubility, our goal is to learn a function $g: \sG \rightarrow \sL$ and predict the labels of other molecular graphs in the testing set. In this paper, the student GNN encoder $f^{S}$ of the learned function $g$ is supervised by our proposed pre-trained teacher model $f^{T}$, which transfers cross-modal knowledge from chemical reactions.

\section{MolKD: The Proposed Method}
\label{sec:method}
In this section, we present our MolKD method for distilling cross-modal knowledge from chemical reactions into molecules.
We first show how to encode structural molecular embeddings in Sec~\ref{sec:encoder}.
After that, we propose the reaction-to-molecule distillation model to transfer cross-modal knowledge from pre-training reaction data into the downstream molecule data in Sec.~\ref{sec:fine-tune}.
Finally, we show how to preserve the equivalence of chemical reactions and incorporate information of yields when pre-training on reactions in Sec.~\ref{sec:pre-train}. 
The workflow of our proposed MolKD is shown in Fig.~\ref{fig:framework}.
Note that for the sake of intuitive representation, we use the condensation reaction of ethyl acetate $\left(\ce{CH3COOH + C2H5OH -> CH3COOC2H5 + H2O}, 0.9\right)$ \footnote{Chemical reactions usually occur under some specific conditions, such as catalysts. The complete equation of this chemical reaction is $\left(\ce{CH3COOH + C2H5OH ->[H2SO4][\Delta] CH3COOC2H5 + H2O}, 0.9\right)$. Following previous work~\cite{wang2021chemical}, we omit the chemical conditions for clarity because they do not affect the conservation law~(\eg, atom number and types) between reactants and products.} as an illustrative example for illustration in what follows.

\subsection{Encoding Structural Molecular Embeddings} 
\label{sec:encoder}
First, we obtain molecular representations in this step.
Intuitively, we take ethanol \ce{C2H5OH} as input, and we get its representation vector $h_{\ce{C2H5OH}}$.
In order to encode rich structure information of the given molecule, we first convert molecular SMILES~\cite{weininger1988smiles} strings to molecular graphs~\footnote{For example, the molecular formula and the SMILES string~(Simplified Molecular-Input Line-Entry System) of ethyl acetate are \ce{C4H8O2} and \ce{CCOC(=O)C}, respectively.}.
We utilize PySmiles~\cite{landrum2013rdkit} library to produce six types of atom properties: \emph{element type}, \emph{(anti)-aromaticness}, \emph{mass}, \emph{the number of implicit hydrogens}, \emph{charge}, and \emph{class}. Each type of atomic property is represented as a one-hot vector. The initial node feature of each atom in the molecular graph is the concatenation of six one-hot vectors.
We then adopt GNN models~(\ie, TAG~\cite{du2017topology}) to encode molecular graphs as numeric vectors. 
\cite{zhang2021mg} argued that hydrogen atoms can help determine the number of chemical bonds for atoms, which are critical to infer the types of atoms especially without taking bond type as input. 
Therefore we regard hydrogen atoms as independent nodes in molecular graphs.
Following~\cite{wang2021chemical}, we do not explicitly encode edge attributes because they can be implicitly inferred by the node features of their own corresponding atoms.

\subsection{Reaction-to-Molecule Distillation Model}
\label{sec:fine-tune}
Second, as shown in Fig.~\ref{fig:framework}(a), we train a molecular property student network supervised by the pre-trained teacher network, given molecule data with property labels. To narrow the gap in data modality between reactions and molecules, we herein introduce the feature-based knowledge distillation method to transfer cross-modal knowledge from a pre-trained teacher network learning with one modality (\ie, reactions) into a student network learning with another modality (\ie, molecules).
Here, we use the toxicity of ethanol \ce{C2H5OH} as the representative molecular property prediction task for illustration.
Intuitively, we fine-tune the student model $f^{S}$ to predict the toxicity of \ce{C2H5OH} supervised by the teacher model $f^{T}$ pre-trained on the reaction $(h_{\ce{CH3COOH}} + h_{\ce{C2H5OH}} \approx h_{\ce{CH3COOC2H5}} + h_{\ce{H2O}}, 0.9)$.

We adopt the feature-based distillation method, \ie, contrastive representation distillation~\cite{tian2019contrastive} and formulate representation distillation as a contrastive learning task on pairwise relationships between the teacher and student molecular representations. Intuitively, we want to maximize the consistency of representations for the teacher and student model. 
Given a minibatch data of molecular property prediction $\gB_1 = \left\{(\gG_1, l_1), (\gG_2, l_2), \cdots \right\}$, we maximize the similarity among pairs of student and teacher representations corresponding to the same molecular graph, \ie, $f^{S}(\gG_i), f^{T}(\gG_i)$~(positive pairs), while pushing away the representations of pairs of unmatched molecules, \ie, $f^{S}(\gG_i), f^{T}(\gG_j)$~(negative pairs). We utilize the InfoNCE loss~\cite{oord2018representation} as the contrastive loss in reaction-to-molecule distillation. Given two molecular encoders $f^{S}$ and $f^{T}$ with the same output feature dimension, we denote representation distillation as the task of classifying positive pairs among a set of negative pairs as
\begin{footnotesize}
\begin{equation}
\begin{aligned}
    &\mathcal{L}_{\text{KD}} (f^{S}, f^{T}) =\\ &-\frac{1}{|\gB_1|}\sum_{i}\log \frac{\ve^{ s\left(f^{S}(\gG_i), f^{T}(\gG_i)\right) / \tau}}{\ve^{s\left(f^{S}(\gG_i), f^{T}(\gG_i)\right) / \tau}+\sum_{j \neq i} \ve^{s\left(f^{S}(\gG_i), f^{T}(\gG_j)\right) / \tau}},
    \label{equ:loss2}
\end{aligned}
\end{equation}
\end{footnotesize}
where $\tau$ represents the temperature hyper-parameter and $s(\cdot)$ indicates the similarity between molecular representations. Here we use the cosine similarity function. The illustration of the distillation loss is shown in Fig.~\ref{fig:framework}(c).

\paragraph{Training Objective.} In MolKD, we adopt TAG~\cite{du2017topology} as our backbone GNN model. 
The workflow of MolKD is shown in Fig.~\ref{fig:framework}. We need to optimize two components of the loss function. First, we have an auxiliary reaction-to-molecule cross-modal distillation loss to obtain a powerful GNN encoder for effective molecular representations~(Eq.~\ref{eq:6}). Second, given the training set of molecule data $\left(\gG,l\right) \in \sG \times \sL$, the learned predictor $g$ consisting of the student GNN encoder $f^{S}$ followed by a classifier is further optimized with a supervised loss:
\begin{equation}
    \mathcal{L}_{\text {sup}} = \mathbb{E}_{\left(\gG,l\right) \in \sG \times \sL} [\mathcal{L}\left(g\left(\gG\right), l\right)] \;.
    \label{equ:loss1}
\end{equation}
Overall, the final training objective of MolKD is
\begin{equation}
    \mathcal{L}=\beta \cdot \mathcal{L}_{\text {sup }}+(1-\beta) \cdot \mathcal{L}_{\text{KD}}\left(f^{S}, \Phi(f^{T})\right),
    \label{equ:loss-all}
\end{equation}
where $\beta$ is a scaling factor to balance the supervised loss and the representation distillation loss, and $\Phi$ is an MLP layer to transform feature maps of the teacher model $f^{T}$ into the same shape as the student model $f^{S}$.

\subsection{Yield-Guided Chemical-Reaction-Aware Pre-Training Model}
\label{sec:pre-train}
Given chemical reactions, we design the dedicated self-supervised pre-training task to measure transformation efficiency of chemical reactions using yields, resulting in an informative pre-trained teacher model.

\paragraph{Preserving Chemical Reaction Equivalence.}
Third, as shown in Fig~\ref{fig:framework}(b), we pre-train on chemical reaction data in this step.
Intuitively, in the molecular embedding space, we obtain $h_{\ce{CH3COOH}} + h_{\ce{C2H5OH}} \approx h_{\ce{CH3COOC2H5}} + h_{\ce{H2O}}$ for the reaction $\ce{CH3COOH + C2H5OH -> CH3COOC2H5 + H2O}$. 
A chemical reaction defines a transformation relation ``$\rightarrow$'' between the reactant set $R = \{ r_1, r_2, \cdots, r_n\}$ and the product set $P = \{ p_1, p_2, \cdots, p_m\}$: 
\begin{equation}
    r_1 + r_2 + \cdots + r_n \longrightarrow p_1 + p_2 + \cdots + p_m .
\end{equation}
A chemical reaction represents a particular relation of \emph{equivalence} between its reactants and products in terms of the conservation of mass and charge. We adopt the idea of preserving the equivalence of chemical reactions in molecular embedding space~\cite{wang2021chemical}. 
This \emph{equivalence} property is reminiscent of the translation-based model in KG~(\eg, TransE) that learns embeddings by
narrowing the distance between the head and tail entity transformed by the relation to preserve its composable property. 
If we take reactants of a reaction as the head entity and products as the tail entity, we can impose constraints on molecular embeddings to preserve the \emph{equivalence} property of the corresponding reaction:
\begin{equation}
    \sum_{r\in R} h_r = \sum_{p\in P} h_p .
\end{equation}

\paragraph{Modeling Transformation Efficiency with Reaction Yields.}
Finally, we pre-train on reactions considering their yields in this step. Intuitively, in the molecule embedding space, we obtain $h_{\ce{CH3COOH}} + h_{\ce{C2H5OH}} \approx h_{\ce{CH3COOC2H5}} + h_{\ce{H2O}}$ for $(\ce{CH3COOH + C2H5OH -> CH3COOC2H5 + H2O}, 0.9)$ considering its reaction yield $0.9$.
The yield is a crucial measure of transformation efficiency of the reactant-product pair in a reaction. 
Intuitively, the higher yield of a reaction suggests that given its reactants, it's more likely to generate the corresponding products than others, \ie, there is higher probability that the relation between the reactants and products is true. This draws a conceptual analogy to the confidence score on an uncertainty KG.
\begin{wrapfigure}[16]{r}{0.25\textwidth}
    \centering
    \includegraphics[width=1\linewidth]{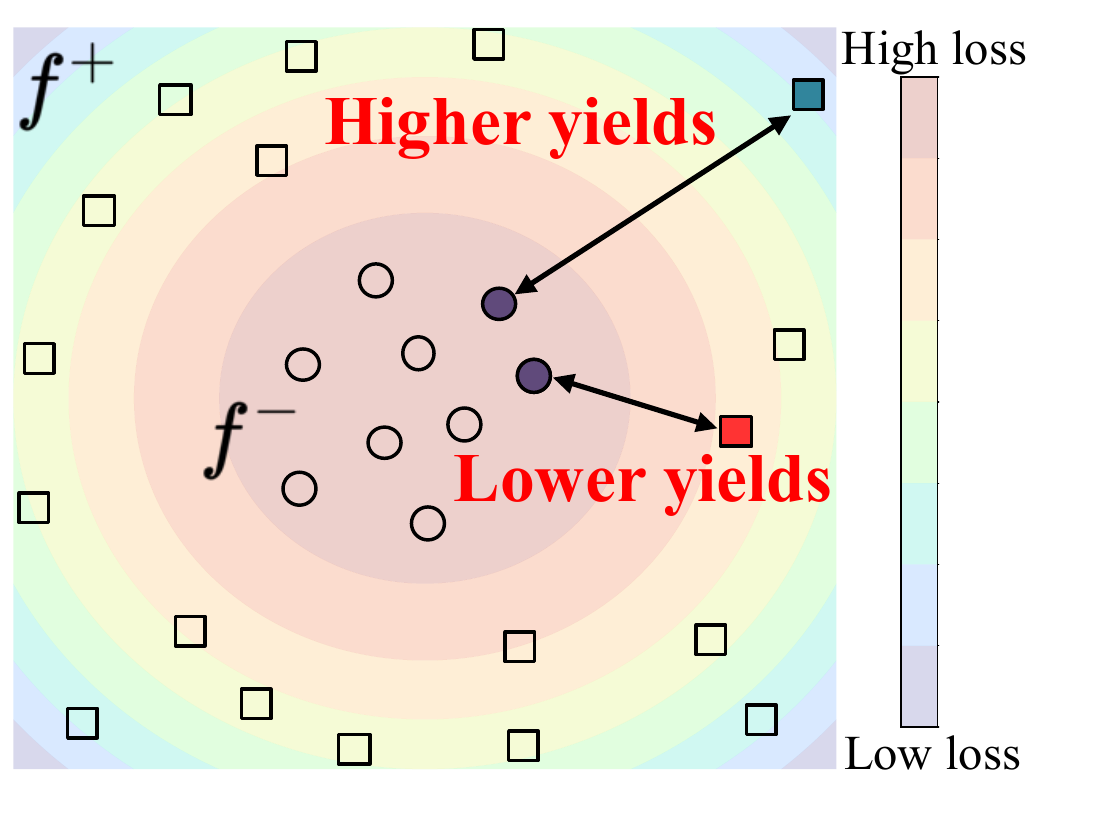}
    \vspace{-1.5em}
    \caption{The illustration of the embedding space with different margins according to reaction yields. Rectangles and circles denote positive and negative samples.}
    \label{fig:illustration}
\end{wrapfigure}
Accordingly, we adopt the idea of GTransE~\cite{kertkeidkachorn2019gtranse} to capture the confidence of reaction triples in an uncertainty KG, by adaptively modifying the margin with the reaction yield in pre-training.
Intuitively, the higher the yield, the larger margin should be set to pay more attention to this reaction. 
Thus, the margin $\gamma$ should be increased when the reaction yield $y$ gets higher.
Specifically, given a minibatch of reaction data $\gB_2=\{(R_1, P_1, y_1), (R_2, P_2, y_2), \cdots\}$, 
we compute a score function to measure the fitness of the reactant-product pair in a reaction. 
We adopt the $L$-$2$ norm introduced in  TransE~\cite{bordes2013translating} to compute the score function with $f(R,P) = \left\|\sum_{r \in R} h_{r}-\sum_{p \in P} h_{p}\right\|_{2}$.
In Fig.~\ref{fig:illustration}, we maximize the margin between the positive sample $f^{+}(R_i, P_i) = \left\|\sum_{r \in R_i} h_{r}-\sum_{p \in P_i} h_{p}\right\|_{2}$ and the negative sample $f^{-}(R_i, P_j) = \left\|\sum_{r \in R_i} h_{r}-\sum_{p \in P_j} h_{p}\right\|_{2}$ of the higher yield~(the blue rectangle) rather than the lower yield one~(the red rectangle).
Formally, if we have $(R_1, P_1, y_1)$, $(R_2, P_2, y_2)$, and $y_1 > y_2$, then $f^{+}(R_1, P_1) - f^{-}(R_1, P_2) > f^{+}(R_2, P_2) - f^{-}(R_2, P_1)$. 
To prevent the GTransE model from learning a trivial solution where all molecular representations are equal to zero, we use the contrastive learning strategy~\cite{chen2020simpleframework} by drawing unpaired reactants and products in a minibatch as negative samples.
Therefore, the margin-loss function of the yield-guided chemical-reaction-aware pre-training model can be denoted as follows:~\footnote{Here we only replace the corrupted products $P$ in negative samples because reaction yields reflect transformation efficiency of the main product in a reaction.}
\begin{footnotesize}
\begin{equation}
\begin{aligned}
    \mathcal{L}_{\mathcal{B}}&=\frac{1}{|\mathcal{B}_2|(|\mathcal{B}_2|-1)} \sum_{i}\sum_{i \neq j} \max \Big(\left\|\sum_{r \in R_{i}} h_{r}-\sum_{p \in P_{i}} h_{p}\right\|_{2}\\ &-\left\|\sum_{r \in R_{i}} h_{r}-\sum_{p \in P_{j}} h_{p}\right\|_{2} + y_i^{\alpha} \gamma, 0\Big),
    \label{eq:6}
\end{aligned}
\end{equation}
\end{footnotesize}
where $\gamma > 0$ is the margin hyper-parameter, and $\alpha \geq 0$ is the hyper-parameter to control the influence of reaction yields. As $\alpha$ becomes larger, the effect of yields on model transformation efficiency is amplified. When $\alpha$ equals zero, the influence of uncertainty is eliminated, and GTransE becomes the traditional translation-based TransE model.

\section{Experiments}
\label{sec:experiments}
In this section, we conduct extensive experiments to answer the following questions: (1) Can our proposed yield-guided chemical-reaction-aware pre-training model provide effective molecular representations?~(Sec.~\ref{sec:exp-pre-train}) (2) How does MolKD compare to several competitive methods on molecular property prediction tasks?~(Sec.~\ref{sec-exp-task}) (3) Can MolKD serve as a reliable method with high robustness and interpretability?~(Sec.~\ref{sec-robust}) 
Each result in this section is obtained by running over 10 runs, and we show the mean results with standard deviation on the testing set. 
Due to space limitations, the descriptions and statistics of the datasets we use are provided in Appendix~\ref{sec:appendix-A}.
Additional empirical studies such as ablations of the backbone GNN models and knowledge distillation methods, as well as the results of regression tasks on molecular property prediction tasks, can be found in Appendix~\ref{sec:appendix-B}.

\begin{table*}[t]
\centering
\small
\caption{Results of chemical reaction prediction on USPTO-500-MT. We use the pre-trained models of Mol2Vec and MolBERT from their respective papers. MolR and MolKD variants are pre-trained on the training set of USPTO-500-MT. 
Arrows indicate the direction of better performance.
\textbf{Bold} / \underline{underline} denote the best / second-best result for each column.}
\label{table:pre-train}
\setlength{\tabcolsep}{19pt}
\scalebox{0.75}{
\begin{tabular}{ccccccc}
\toprule
                 & MRR $\uparrow$ & MR $\downarrow$   & Hit@1 $\uparrow$       & Hit@3 $\uparrow$         & Hit@5 $\uparrow$         & Hit@10 $\uparrow$        \\ \midrule
Mol2Vec          & 0.758                     & 93.748          & 0.696        & 0.799        & 0.830        & 0.864         \\
MolBERT          & 0.796                     & 18.781          & 0.720        & 0.853        & 0.888        & 0.923                \\
MolR             & $0.853_{\pm0.005}$              & $2.351_{\pm0.108}$    & $0.783_{\pm0.006}$ & $0.911_{\pm0.005}$ & $0.941_{\pm0.004}$ & $0.968_{\pm0.002}$  \\ \midrule
MolKD-random     & $0.725_{\pm0.017}$              & $174.740_{\pm49.859}$ & $0.669_{\pm0.021}$ & $0.762_{\pm0.015}$ & $0.790_{\pm0.013}$ & $0.822_{\pm0.011}$ \\
MolKD-confidence & \underline{$0.858_{\pm0.005}$}              & $2.271_{\pm0.106}$    & \underline{$0.789_{\pm0.007}$} & \underline{$0.915_{\pm0.005}$} & \underline{$0.945_{\pm0.004}$} & \underline{$0.970_{\pm0.002}$}  \\
MolKD-CKRL       & $0.857_{\pm0.005}$              & \underline{$2.260_{\pm0.099}$}    & $0.787_{\pm0.007}$ & $0.913_{\pm0.005}$ & $0.944_{\pm0.003}$ & \underline{$0.970_{\pm0.002}$}  \\
MolKD            & \bm{$0.875_{\pm0.003}$}              & \bm{$2.058_{\pm0.054}$}    & \bm{$0.811_{\pm0.003}$} & \bm{$0.929_{\pm0.002}$} & \bm{$0.955_{\pm0.002}$} & \bm{$0.977_{\pm0.002}$} \\ \bottomrule
\end{tabular}
}
\end{table*}
\subsection{Pre-Training on Chemical Reactions}
\label{sec:exp-pre-train}
\paragraph{Datasets.} We use USPTO-500-MT collected by~\cite{lu2022unified}, a public large-scale reaction dataset including reaction SMILES strings and their yields, to validate the effectiveness of the yield-guided chemical-reaction-aware pre-training method. This dataset consists of 143,535 reactions: 116,360/12,937/14,238 for training/validation/testing. Each reaction has at most 5 reactants and only 1 main product.
\paragraph{Baselines.} We compare MolKD with various pre-training methods for molecular representations: Mol2vec~\cite{jaeger2018mol2vec}, MolBERT~\cite{fabian2020molecular}, and MolR~\cite{wang2021chemical}. Moreover, we carry out the following ablations: (1) we replace reaction yields with \whiteding{1} random numbers that fall into $[0,1]$~(MolKD-random) \whiteding{2} confidence scores~\cite{ghiandoni2019development} that measure the quality of reaction classes~(MolKD-confidence). These two ablations are designed to highlight the importance of yields when pre-training on reactions. (2) we adopt another uncertainty KG method, \ie, CKRL~\cite{xie2018does}) to validate the versatility of our proposed model~(MolKD-CKRL).
\paragraph{Setups.} Inspired by the equivalence property between reactants and products that $h_{R} \approx h_{P}$, we treat the chemical reaction prediction as a ranking problem following~\cite{wang2021chemical}. In the testing phase, given the reactants of a reaction, we predict its main product and rank all candidate products~(there are 14,123 in total) according to $l$-2 distances in the representation space $||h_R-h_p||_2$. We conduct three measures as our evaluation metrics. (1) MRR: mean reciprocal rank (2) MR: mean rank of correct entities (3) Hit@K: the proportion of correct answers ranked in top K.

Table~\ref{table:pre-train} shows the results of chemical reaction predictions, from which we can observe that: (1) Our MolKD model achieves the best performance compared with other baselines and ablations over 6 evaluation metrics. For example, MolKD achieves $2.2\%$ absolute MRR gain and $2.8\%$ absolute Hit@1 gain compared with the best-performing MolR model, which confirms the capability of our yield-guided pre-training method. (2) Compared with MolR~(without considering yields), MolKD-random~(replacing yields with meaningless numbers), and MolKD-confidence~(replacing yields with reaction confidence scores~\cite{ghiandoni2019development}), MolKD achieves $2.2\%/12.0\%/1.7\%$ absolute MRR improvement on MRR, 
highlighting the effectiveness of introducing the important chemical reaction factor--yields when pre-training molecular representations. (3) MolKD achieves slightly better performance compared with MolKD-CKRL, which suggests that MolKD does not rely on the specific uncertainty KG method to achieve strong performance.

\begin{figure}[]
    \centering
    \includegraphics[width=\linewidth]{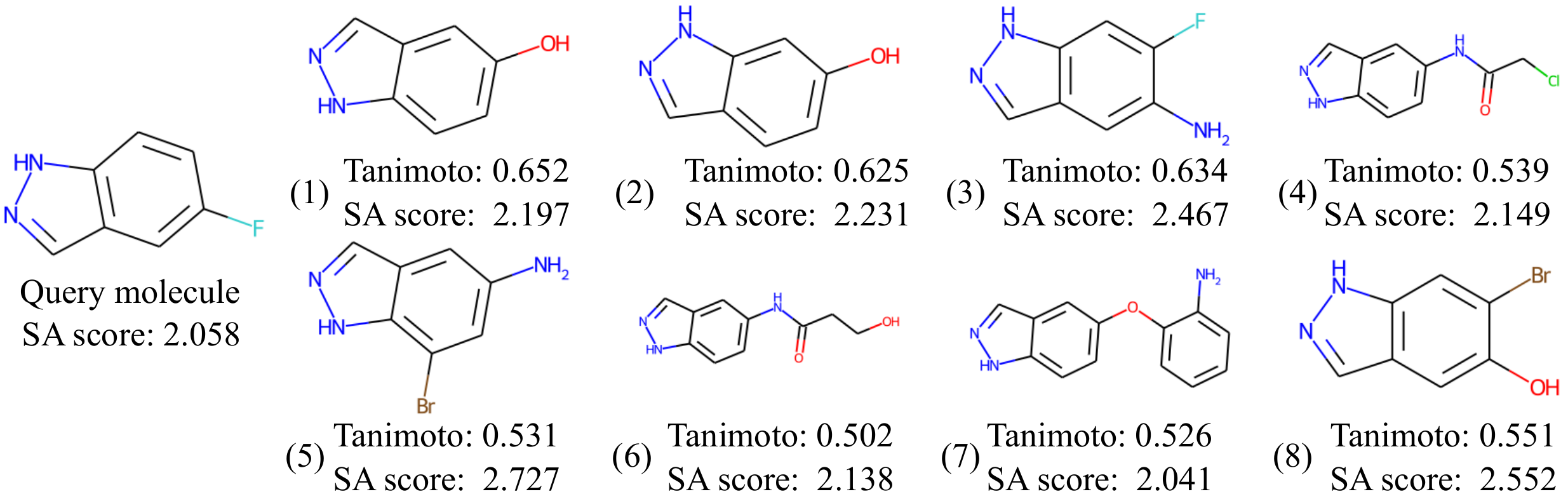}
    \caption{Comparisons of the query molecule~(PubChem ID 17842486) and eight closest molecules in MolKD representation domain with Tanimoto similarities and SA score labeled.}
    \vspace{-1em}
    \label{fig:molecule-query}
\end{figure}

\paragraph{Investigation of Pre-Training Representations.} To examine the effectiveness of the representations learned by our pre-trained method in MolKD, we visualize a representative molecule with its eight closest molecules in the representation domain. Specifically, given the query molecule~(PubChem ID 17842486), we obtain its representations via the yield-guided pre-training method in MolKD and calculate cosine distances with all reference molecules in our pre-training dataset. The cosine distance between two molecular representations $(\vu, \vv)$ is defined as $1-\frac{\vu \cdot \vv}{||\vu|| ||\vv||}$. Then, all reference molecules are ranked by cosine distances.
In Fig.~\ref{fig:molecule-query}, we show the eight closest molecules compared to the query molecule.
We also calculate molecular similarities through Tanimoto coefﬁcient~\cite{bajusz2015tanimoto} between the query and the selected molecule, and SA scores~\cite{ertl2009estimation} to assess the associated molecule's synthetic complexity.
We find that these molecules are structurally similar to the query molecule with the same functional groups~(\eg, aromatics).
They have high Tanimoto similarities larger than 0.5. The learned representations via MolKD are in line with our expectation that molecules with similar structures tend to be close in the representation domain. Moreover, these molecules have similar SA scores, which further suggests that selected molecules share similar synthetic complexities in absence of the activity cliff issue~\cite{hu2012extending}.


\begin{table*}[]
\centering
\small
\caption{Results of molecular property prediction on classification tasks~(metric: AUC-ROC, higher is better). The results of other models in the random and the scaffold split are taken from \protect\cite{wang2021chemical} and \protect\cite{xia2023molebert}, respectively. ``-'' means they do not report the corresponding results.}
\label{table:classification-new}
\setlength{\tabcolsep}{13pt}
\scalebox{0.8}{
\begin{tabular}{c|ccccccc}
\toprule Split & Method                  & BACE                                  & BBBP                                  & ClinTox                               & HIV                                   & SIDER                                 & Tox21                                 \\ \midrule
\multirow{11}{*}{Random} & AttentiveFP                         & 0.867                                 & 0.729                                 & 0.799                                 & 0.792                                 &                    -                   & 0.822                                 \\
& GraphConv                           & 0.783                                 & 0.69                                  & 0.807                                 & 0.763                                 & 0.638                                 & 0.829                                 \\
& Weave                               & 0.806                                 & 0.671                                 & 0.832                                 & 0.703                                 & 0.621                                 & 0.82                                  \\
& MPNN                                & 0.815                                 &                   -                    & 0.879                                 &                   -                    & 0.641                                 & 0.808                                 \\
& Mol2vec                             & $0.862_{\pm0.027}$                          & $0.872_{\pm0.021}$                          & $0.841_{\pm0.062}$                          & $0.769_{\pm0.021}$                          &               -                        & $0.803_{\pm0.041}$                          \\
& ChemBERTa                           &           -                            & 0.643                                 & 0.733                                 & 0.622                                 &                    -                   & 0.728                                 \\
& MolBert                             & 0.866                                 & 0.762                                 &                         -              & 0.783                                 &               -                        & 0.806                                 \\
& MolR-GCN                            & $\bm{0.882_{\pm0.019}}$                          & $0.890_{\pm0.032}$                          & $0.916_{\pm0.039}$                          & $0.802_{\pm0.024}$                          &                    -                   & $0.818_{\pm0.023}$                          \\
& MolR-TAG                            & $0.875_{\pm0.023}$   & $0.895_{\pm0.031}$   & $0.913_{\pm0.043}$   & $0.801_{\pm0.023}$   &       -         & $\underline{0.820_{\pm0.028}}$   \\
& MolKD$^{-}$                               & $\underline{0.876_{\pm0.031}}$  & $\underline{0.902_{\pm0.028}}$  & $\underline{0.917_{\pm0.039}}$ & $\underline{0.811_{\pm0.021}}$ & $\underline{0.660_{\pm0.048}}$   & $0.818_{\pm0.059}$ \\
& MolKD (ours)                        & $0.872_{\pm0.022}$ & $\bm{0.915_{\pm0.021}}$ & $\bm{0.933_{\pm0.041}}$ & $\bm{0.816_{\pm0.011}}$ & $\bm{0.706_{\pm0.020}}$ & $\bm{0.841_{\pm0.032}}$ \\ \midrule\midrule
\multirow{13}{*}{Scaffold} & PretrainGNN(GIN)                    & $0.778_{\pm0.018}$                             & $0.652_{\pm0.014}$                             & $0.735_{\pm0.043}$                             & $0.753_{\pm0.015}$                             & $0.605_{\pm0.009}$                             & $0.751_{\pm0.009}$                             \\
& InfoGraph                           & $0.743_{\pm0.026}$                             & $0.687_{\pm0.006}$                             & $0.754_{\pm0.043}$                             & $0.742_{\pm0.009}$                             & $0.587_{\pm0.006}$                             & $0.733_{\pm0.006}$                             \\
& GraphLOG                            & $0.786_{\pm0.010}$                             & $0.683_{\pm0.016}$                             & $0.617_{\pm0.026}$                             & $0.737_{\pm0.011}$                             & $0.576_{\pm0.019}$                             & $0.734_{\pm0.006}$                              \\
& G-Contextual                        & $\underline{0.793_{\pm0.011}}$                             & $0.699_{\pm0.021}$                             & $0.606_{\pm0.052}$                             & $\underline{0.763_{\pm0.015}}$                             & $0.587_{\pm0.010}$                             & $0.750_{\pm0.006}$                             \\
& G-Motif                             & $0.730_{\pm0.033}$                             & $0.669_{\pm0.031}$                             & $0.777_{\pm0.027}$                             & $0.738_{\pm0.012}$                             & $0.610_{\pm0.015}$                            & $0.736_{\pm0.007}$                             \\
& KCL                                 & $0.742_{\pm0.013}$                             & $0.653_{\pm0.010}$                             & $0.653_{\pm0.052}$                             & $0.755_{\pm0.007}$                             & $0.595_{\pm0.007}$                             & $0.742_{\pm0.006}$                             \\
& JOAO                                & $0.732_{\pm0.016}$                             & $0.664_{\pm0.010}$                             & $0.666_{\pm0.031}$                             & $\bm{0.769_{\pm0.007}}$                             & $0.604_{\pm0.015}$                             & $0.748_{\pm0.006}$                             \\
& 3D InfoMax                          & $0.786_{\pm0.019}$                             & $0.691_{\pm0.012}$                             & $0.627_{\pm0.033}$                             & $0.761_{\pm0.013}$                             & $0.568_{\pm0.021}$                             & $0.745_{\pm0.007}$                             \\
& GraphMAE                            & $0.762_{\pm0.019}$                             & $\underline{0.712_{\pm0.010}}$                             & $0.735_{\pm0.030}$                             & $0.758_{\pm0.006}$                             & $0.605_{\pm0.012}$                             & $\underline{0.752_{\pm0.009}}$                             \\
& SimGRACE                            & $0.749_{\pm0.020}$                             & $\underline{0.712_{\pm0.011}}$                             & $0.755_{\pm0.020}$                             & $0.750_{\pm0.006}$                             & $0.602_{\pm0.009}$                             & $0.744_{\pm0.003}$                             \\
& GraphMVP                            & $0.768_{\pm0.011}$                             & $0.685_{\pm0.002}$                             & $0.790_{\pm0.025}$                             & $0.748_{\pm0.014}$                             & $\bm{0.623_{\pm0.016}}$                             & $0.745_{\pm0.004}$                           \\
& MolKD$^{-}$       & $0.772_{\pm0.001}$  & $0.682_{\pm0.001}$ & $\underline{0.805_{\pm0.001}}$ & $0.733_{\pm0.002}$ & $0.603_{\pm0.001}$ & $0.742_{\pm0.035}$ \\
& MolKD (ours) & $\bm{0.801_{\pm0.008}}$ & $\bm{0.748_{\pm0.023}}$ & $\bm{0.838_{\pm0.031}}$  & $0.749_{\pm0.017}$  & $\underline{0.613_{\pm0.012}}$   & $\bm{0.774_{\pm0.024}}$ \\ \bottomrule
\end{tabular}
}
\end{table*}
\subsection{Molecular Property Prediction}
\label{sec-exp-task}
\paragraph{Datasets.} We benchmark the performance of MolKD on multiple challenging classification and regression datasets curated from MoleculeNet~\cite{wu2018moleculenet}. Following~\cite{wang2021chemical,fang2022geometry}, all datasets are randomly split into the training, validation, and testing set by 8:1:1. We use BACE, BBBP, ClinTox, HIV, SIDER, and Tox21 for classification tasks and adopt AUC-ROC as the evaluation metric for these binary prediction tasks, for which higher is better. 
We use two split types for classification tasks: random split~\cite{wang2021chemical} and scaffold split~\cite{xia2023molebert} to thoroughly evaluate the effectiveness of MolKD. 
The scaffold splitting method splits molecules according to their molecular substructure, which is a more challenging splitting method and can better evaluate the generalization ability of models on out-of-distribution data samples.
We use ESOL, FreeSolv, Lipo, and QM8 for regression tasks.
Following~\cite{liu2022structured,li2022glam}, we adopt the root mean square error~(RMSE) for ESOL, Lipo, and FreeSolv, whereas we adopt mean average error~(MAE) for QM8, for which lower is better.
We use the scaffold split for all regression tasks.
\paragraph{Baselines.} For random split, we compare MolKD with multiple competitive baselines. AttentiveFP~\cite{rogers2010extended}, GraphConv~\cite{kipf2016semi}, Weave~\cite{rogers2010extended}, and MPNN~\cite{huang2021therapeutics} are the GNN-based methods without pre-training.
Mol2vec~\cite{jaeger2018mol2vec}, ChemBERTa~\cite{chithrananda2020chemberta}, MolBert~\cite{fabian2020molecular}, and MolR~\cite{wang2022molecular} are molecular pre-training methods. 
For scaffold split, we compare MolKD with the following competitive pre-training GNN baselines: PretrainGNN~\cite{hu2019strategies}, InfoGraph~\cite{sun2019infograph}, GraphLOG~\cite{xu2021self}, G-Contextual/G-Motif~\cite{rong2020self}, KCL~\cite{fang2022molecular}, JOAO~\cite{you2021graph}, 3D infomax~\cite{stark20223d}, GraphMAE~\cite{hou2022graphmae}, SimGRACE~\cite{xia2022simgrace}, and GraphMVP~\cite{liu2022structured}.
MolKD$^-$ means we directly use the pre-trained GNN model followed by a logistic regression layer to make predictions, which does not use the reaction-to-molecule distillation model.

The overall performance of MolKD along with other baselines on classification tasks are summarized in Table~\ref{table:classification-new}. 
We have the following observations: 
(1) On the random split, MolKD achieves the state-of-the-art performance over 5 out of 6 classification tasks, which demonstrates the effectiveness of two main components in MolKD: the yield-guided pre-training model and the reaction-to-molecule distillation model. For example, MolKD achieves $2.1\%$ absolute AUC-ROC gain on Tox21 compared with the best-performing method MolKD-TAG. 
(2) On the scaffold split, MolKD outperforms several competitive pre-training methods over 4 out of 6 datasets. Scaffold splitting method simulates the real-world use case and verifies the out-of-distribution generalization ability of MolKD. For example, MolKD achieves $2.2\%$ absolute AUC-ROC gain on Tox21 compared to the second best model.
(3) Methods with pre-training on large-scale unlabeled molecules consistently outperform methods without pre-training. These results highlight the importance of introducing auxiliary data sources, \eg, chemical reactions for molecular representations.

\subsection{Experimental Analysis}
\begin{table}[]
\centering
\small
\caption{Effect score of molecular structure perturbation test.}
\label{table:robust}
\setlength{\tabcolsep}{13pt}
\scalebox{0.85}{
\begin{tabular}{cccc}
\toprule
\multirow{2}{*}{Method} & \multicolumn{3}{c}{Effect score$_{std}$ {[}lower is better{]}} \\ \cmidrule{2-4} 
       & Level1              & Level2              & Level3             \\ \midrule
GCN    & $0.385_{\pm0.161}$       & $0.712_{\pm0.169}$       & $0.997_{\pm0.183}$      \\
GAT    & $0.388_{\pm0.055}$       & $0.615_{\pm0.087}$       & $0.943_{\pm0.145}$      \\
GIN    & $0.312_{\pm0.017}$       & $0.526_{\pm0.039}$       & $0.764_{\pm0.015}$      \\
MPNN   & $0.315_{\pm0.014}$       & $0.518_{\pm0.054}$       & $0.750_{\pm0.048}$      \\
GLAM   & \underline{$0.290_{\pm0.010}$}       & \underline{$0.493_{\pm0.074}$}       & \bm{$0.656_{\pm0.118}$}      \\
MolKD  & \bm{$0.251_{\pm0.021}$}      & \bm{$0.491_{\pm0.036}$}      & \underline{$0.710_{\pm0.035}$}     \\ \bottomrule
\end{tabular}
}
\end{table}
\label{sec-robust}
\paragraph{Robustness against Molecular Structure Perturbation.} To evaluate the robustness of our proposed method, we follow the principle of property-slightly-affected structure perturbation~(PASP) introduced by~\cite{li2022glam} and utilize the PhysProp dataset to perform a robustness experiment. 
The aim is to investigate that the model does not significantly affect the predictions~(molecular properties) when the input perturbed molecule set suffers small perturbations.
More details of the evaluation metric~(\ie, effect score) are provided in Appendix~\ref{sec:appendix-A}. 
Table~\ref{table:robust} shows that MolKD achieves the best performance with high robustness on the level 1\&2 perturbations and is less affected by molecular structure perturbations.
We postulate the reason is that pre-training on reactions could incorporate comprehensive chemical domain knowledge to increase the robustness of the predictor. 

\begin{figure}[t]
    \centering
    \includegraphics[width=0.8\linewidth]{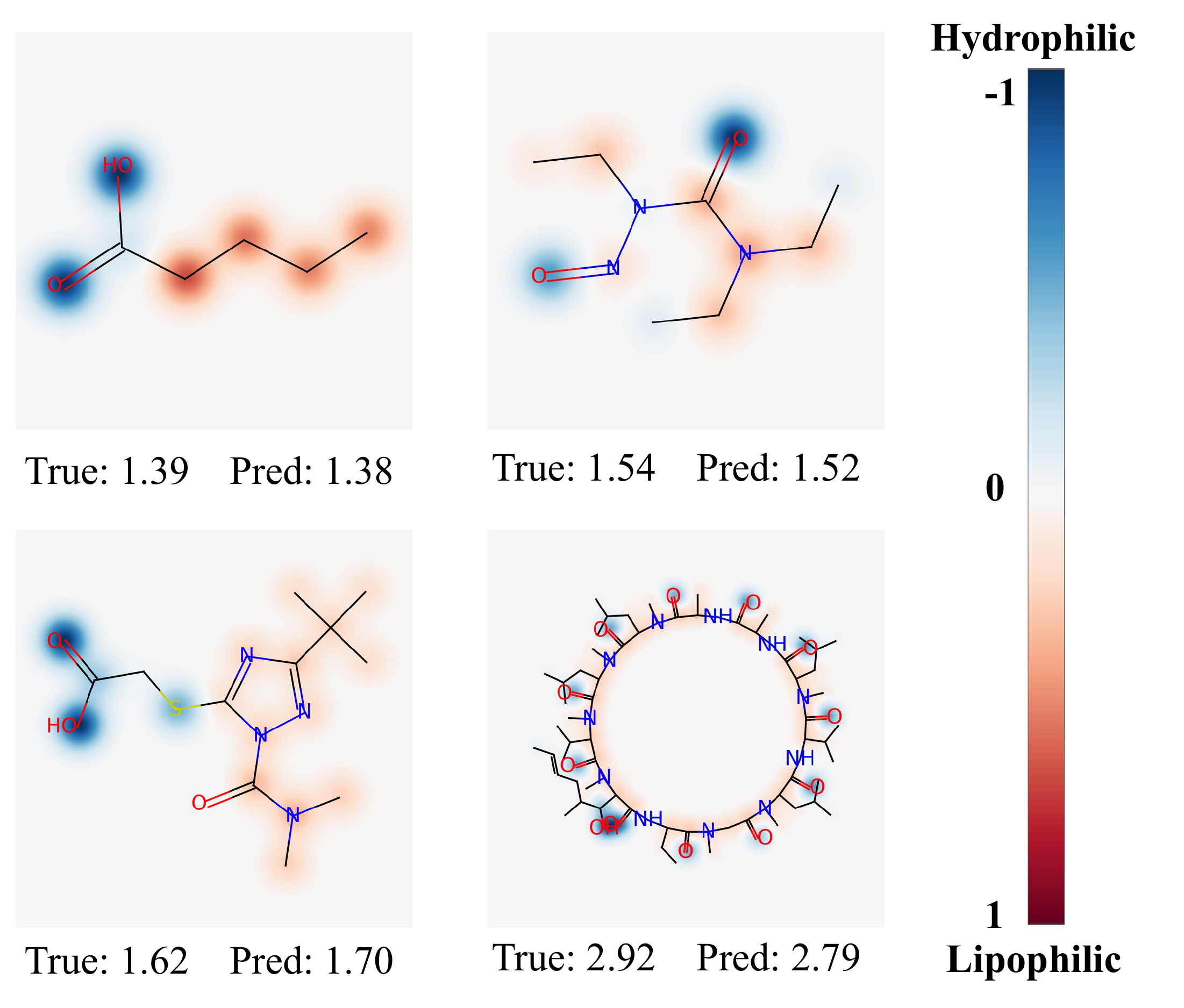}
    \caption{Case studies of atom-level interpretation with true and predicted molecular solubility labeled.}
    \vspace{-1em}
    \label{fig:lipo}
\end{figure}
\paragraph{Interpretability.}
To better understand the predictors generated by MolKD, we investigate its decision-making process and interpret its learned knowledge on PhysProp~\cite{li2022glam}. As shown in Fig.~\ref{fig:lipo}, we visualize some case studies of the solubility prediction and explain the model from the hidden embeddings of the last layer in the GNN model by averaging scaling. 
In general, hydroxyl and amino groups are considered to be more hydrophilic, and alkyl and halogen groups are considered to be more lipophilic. 
In Fig.~\ref{fig:lipo}, the atoms in the hydrophilic group~(\eg, -OH, and -COOH) tend to be bluer and their weights are closer to $1$ in our visualization. Meanwhile, the atoms in the lipophilic group~(\eg, benzene ring) tend to be redder and their weights are closer to $-1$. 
These results are in line with the chemical intuition, which suggests that MolKD captures interpretable knowledge and patterns in chemistry. The full version of case studies of interpretability can be found in Appendix~\ref{sec:appendix-B}.


\section{Conclusion}
\label{sec:conclusion}
In this work, we study how to distill cross-modal knowledge in chemical reactions to assist molecular property prediction.
We propose a novel method named MolKD to obtain effective molecular representations, which consists of two main components: the reaction-to-molecule knowledge distillation model and the yield-guided chemical-reaction-aware pre-training model.
Through extensive experiments on chemical reaction prediction and molecular property prediction, we show that MolKD achieves significantly superior performance with high robustness and interpretability.
We hope our work could stimulate more ideas to squeeze the potential of chemical domain knowledge for molecular property prediction.

\bibliographystyle{ijcai23}
\bibliography{ijcai23-simple}

\clearpage
\appendix
\section{Experimental Details}
\label{sec:appendix-A}
\begin{table*}[h]
\centering
\small
\caption{Statistics of the used datasets.}
\label{table:stat}
\setlength{\tabcolsep}{9pt}
\scalebox{0.9}{
\begin{tabular}{llcccccc}
\toprule
Category                            & Dataset            & Data Type              & Task Type      & \#Molecules & Avg. \#Atoms & Avg. \#Edges & Metric  \\ \midrule
\makecell*[c]{Quantum\\Mechanics}                   & QM8                & \makecell*[c]{SMILES,\\3D coordinates} & Regression     & 21,786       & 15.9         & 24.2         & MAE     \\ \midrule
\multirow{3}{*}{\makecell*[c]{Physical\\Chemistry}} & ESOL               & SMILES                 & Regression     & 1,128        & 13.3         & 13.7         & RMSE    \\
                                    & FreeSolv           & SMILES                 & Regression     & 642         & 8.7          & 8.4          & RMSE    \\
                                    & Lipophilicity      & SMILES                 & Regression     & 4,200        & 27.0         & 29.5         & RMSE    \\ \midrule
\multirow{2}{*}{Biophysics}         & HIV                & SMILES                 & Classification & 41,127       & 25.5         & 27.5         & ROC-AUC \\
                                    & BACE               & SMILES                 & Classification & 1,513        & 34.1         & 36.9         & ROC-AUC \\ \midrule
\multirow{4}{*}{Physiology}         & BBBP               & SMILES                 & Classification & 2,039        & 24.1         & 26.0         & ROC-AUC \\
                                    & Tox21              & SMILES                 & Classification & 7,831        & 18.6         & 19.3         & ROC-AUC \\
                                    & SIDER              & SMILES                 & Classification & 1,427        & 33.6         & 35.4         & ROC-AUC \\
                                    & ClinTox            & SMILES                 & Classification & 1,478        & 26.2         & 27.9         & ROC-AUC \\ \midrule
Robustness                          & \makecell*[l]{Perturbed\\PhysProp} & SMILES                 & Regression     & 12,607       & 16.9         & 26.0         & Effect score    \\ \bottomrule
\end{tabular}
}
\end{table*}
In this section, we first provide the detailed descriptions and statistics of all datasets used in Sec.~\ref{sec:experiments}. We then list hyper-parameters of our proposed MolKD method followed by the experimental settings.

\subsection{Dataset Descriptions and Statistics}
To benchmark the performance of our MolKD model, we use 9 benchmark datasets curated from MoleculeNet~\cite{wu2018moleculenet} including both classification and regression tasks. These datasets cover a wide range of categories of molecular tasks, \ie, quantum mechanics, physical chemistry, biophysics, and physiology. All datasets are randomly split into training, validation, and testing subsets following an 80/10/10 ratio. Furthermore, in order to investigate the robustness of the proposed model, we adopt the perturbed PhysProp dataset to estimate the PASP~(property-slightly-affected structure perturbation) property~\cite{li2022glam}.
Following~\cite{wu2018moleculenet,li2022glam}, different classification and regression metrics are used to fairly compare different methods: \emph{ROC-AUC}~(Area Under Curve of Receiver Operating Characteristics) for classification tasks, \emph{RMSE}~(Root-Mean-Square Error) for regression tasks except for the QM8 dataset~(\ie, ESOL, FreeSolv, Lipo, and QM8), \emph{MAE}~(Mean Absolute Error) for the QM8 dataset, and \emph{Effect score} for the perturbed PhysProp dataset.
The statistics of datasets we used in experiments are summarized in Table~\ref{table:stat}. We also provide a brief description of datasets as follows~\cite{huang2021therapeutics,li2022glam}:
\begin{itemize}[leftmargin=10pt]
    \item \textbf{QM8:} $21,786$ small molecules whose regression labels are electronic spectra and excited state energy calculated by multiple quantum mechanic methods.
    \item \textbf{ESOL:} $1,128$ common organic small molecules whose labels are water solubility~(log solubility in mols per litre).
    \item \textbf{FreeSolv:} $642$ small molecules whose regression labels are experimental and calculated hydration free energy in water.
    \item \textbf{Lipophilicity:} $4,200$ molecules whose regression labels are experimental results of octanol/water distribution coefficient~(logD at pH 7.4).
    \item \textbf{HIV:} $41,127$ molecules whose classification labels are experimentally measured abilities to inhibit HIV replication. 
    \item \textbf{BACE:} $1,513$ molecules whose classification labels are binary binding results for a set of inhibitors of human $\beta$-secretase 1~(BACE-1).
    \item \textbf{BBBP:} $2,039$ molecules with binary labels of blood-brain barrier penetration~(permeability).
    \item \textbf{Tox21:} $7,831$ molecules with qualitative toxicity measurements on the biological target. 
    \item \textbf{SIDER:} $1,427$ molecules which are collected from marketed drugs and adverse drug reactions~(ADR) dataset.
    \item \textbf{ClinTox:} Qualitative data of $1,478$ drugs approved by the FDA and those that have failed clinical trials for toxicity reasons.
    \item \textbf{Perturbed PhysProp:} $14,176$ molecules structures and their corresponding lipophilicity properties~(logP), whose principle is to determine an ideal perturbed molecule set with small perturbations that do not significantly affect the properties. In order to obtain the perturbed data, \cite{li2022glam} compare all possible molecule pairs in PhysProp, calculate the fingerprint similarity of all molecules and their difference in logP, and pick out molecule pairs that meet the following two conditions: (1) the difference in the logP of the molecule pairs should be less than 0.2; (2) the molecular fingerprint similarity should be in the range of 0.3–1.0. These molecule pairs are then divided into three levels~(range 0.8–1.0, 0.5–0.8, and 0.3–0.5 marked as levels 1, 2, and 3). Finally, those molecules that exist in all three levels constitute the perturbed PhysProp dataset. According to~\cite{li2022glam}, the evaluation metric of \emph{effect score} is defined as follows: given a molecule set $M$ with ground-truth properties $Q$ and a trained predictor $f$, we predict the property set $P$ by
    \begin{equation}
        P = f(M)
    \end{equation}
    Similarly, Given the perturbed set $M'$ with properties $Q'$, we predict the property set $P'$ by
    \begin{equation}
        P' = f(M')
    \end{equation}
    Finally, the perturbation effect score $\Delta$ of method $f$ is calculated by
    \begin{equation}
        \Delta = L(P, P') - L(Q, Q'),
    \end{equation}
    where we use r.m.s.e as our distance function $L$.
\end{itemize}

\subsection{Hyper-Parameters}
In the pre-training process of chemical reaction data, we use the following three commonly-used GNNs as the implementation of our molecular encoder: TAG~\cite{du2017topology}, GCN~\cite{kipf2016semi}, and GIN~\cite{xu2018powerful}. We use the default hyper-parameters as introduced in the PyTorch-Geometric library for each GNN model.
The number of propagation layers for all GNN models is $2$, and the output dimension of GNNs is 2,048. 
The margin $\gamma$ and $\alpha$ in Eq.~\ref{eq:6} is set to $6$ and $2$ respectively. 
We train the model for $20$ epochs with a batch size of $4,094$.
We use the Adam~\cite{kingma2014adam} optimizer with a learning rate of $1\times10^{-4}$, $\beta_1$ of $0.9$, and $\beta_2$ of $0.999$.

In the fine-tuning process of molecular property prediction data, the hidden dimension of the student GNN backbone model is $512$, which is followed by two fully-connected layers to output the prediction. 
We carry out a grid search on the validation dataset to find the optimal temperature hyper-parameter $\tau$ in Eq.~\ref{equ:loss2} and $\beta$ in Eq.~\ref{equ:loss-all}.
We tune $\tau=\{0.05, 0.075, 0.1\}$ and $\alpha=\{0.2, 0.5, 0.8\}$.
For example, we set $\tau=0.05$ and $\alpha=0.5$ on the QM8 dataset. We train the model for $400$ epochs with a batch size of $1,024$. The optimization is conducted using Adam~\cite{kingma2014adam} with a learning rate of $2\times10^{-4}$.

\subsection{Experimental Settings}
All experiments are conducted with the following settings:
\begin{itemize}[leftmargin=10pt]
    \item Operating system: Linux Red Hat 4.8.2-16
    \item CPU: Intel(R) Xeon(R) Platinum 8255C CPU @ 2.50GHz
    \item GPU: NVIDIA Tesla V100 SXM2 32GB
    \item Software versions: Python 3.8.10; Pytorch 1.9.0+cu102; Numpy 1.20.3; SciPy 1.7.1; Pandas 1.3.4; Scikit-learn 1.0.1; PyTorch-geometric 2.0.2; DGL 0.7.2; Open Graph Benchmark 1.3.2
\end{itemize}

\section{More Experimental Results}
\label{sec:appendix-B}
In this section, we first present the experimental results of molecular property prediction on regression tasks.
Then, we investigate the choice of GNN backbone models and knowledge distillation methods as ablation studies. Finally, We provide full cases on USPTO-500-MT and PhysProp to illustrate that the proposed method can serve as an effective predictor with high robustness and interpretability.

\subsection{Molecular Property Prediction on Regression Tasks}
\begin{table}[]
\centering
\caption{Results of molecular property prediction on regression tasks~(metric: RMSE for ESOL, FreeSolv, and Lipo; MAE for QM8, lower is better).}
\label{table:regression-new}
\setlength{\tabcolsep}{6pt}
\scalebox{0.73}{
\begin{tabular}{ccccc}
\toprule
                 & ESOL                                  & FreeSolv                             & Lipo                                  & QM8                                     \\ \midrule
SVM              & $1.50_{\pm0.00}$                            & $3.14_{\pm0.00}$                           & $0.82_{\pm0.00}$                            & $0.0543_{\pm0.0010}$                          \\
GCN              & $1.43_{\pm0.05}$                            & $2.87_{\pm0.14}$                           & $0.85_{\pm0.08}$                            & $0.0366_{\pm0.0011}$                          \\
GIN              & $1.45_{\pm0.02}$                            & $2.76_{\pm0.18}$                           & $0.85_{\pm0.07}$                            & $0.0371_{\pm0.0009}$                          \\
SchNet           & $1.05_{\pm0.06}$                            & $3.22_{\pm0.76}$                           & $0.91_{\pm0.10}$                            & $0.0204_{\pm0.0021}$                          \\
PretrainGNN(GIN) & $1.10_{\pm0.01}$                          & $2.76_{\pm0.01}$                         & $\underline{0.74_{\pm0.01}}$                         & $\underline{0.0200_{\pm0.0001}}$                            \\
GROVE-base       & $\underline{0.98_{\pm0.09}}$                           & $\underline{2.18_{\pm0.05}}$                         & $0.82_{\pm0.01}$                          & $0.0218_{\pm0.0004}$                          \\
GROVE-large      & $\bm{0.90_{\pm0.02}}$                          & $2.27_{\pm0.05}$                         & $0.82_{\pm0.01}$                           & $0.0224_{\pm0.0003}$                          \\
JOAO             & $1.12_{\pm0.02}$                          &                     -                 & $\bm{0.71_{\pm0.01}}$                          &               -                          \\
GraphMVP         & $1.09$                                 &                      -                & $\bm{0.71}$                                 &           -                              \\
MolKD$^{-}$           &  $1.05_{\pm0.06}$ &  $2.64_{\pm0.16}$ &  $0.96_{\pm0.03}$ &  $0.0240_{\pm0.0026}$ \\
MolKD (ours)     &  $\bm{0.90_{\pm0.04}}$ &  $\bm{1.94_{\pm0.03}}$ &  $0.89_{\pm0.01}$ &  $\bm{0.0173_{\pm0.0013}}$ \\ \bottomrule
\end{tabular}
}
\end{table}
We further represent the results of molecular property prediction on regression tasks.
As shown in Table~\ref{table:regression-new}, MolKD consistently outperforms other competitive baseline models on three regression tasks in molecular property prediction.
For example, MolKD achieves [$0.82, 0.24$] absolute RMSE gain on FreeSolv compared with the best-performing method [AttentiveFP~(w/o pre-training), GROVE-base~(with pre-training)], and [$15.20\%, 13.50\%$] MAE gain on QM8 compared with the best-performing method [SchNet~(w/o pre-training), PretrainGNN~(with pre-training)]. These results further highlight the effectiveness of our proposed model.

\subsection{Ablation Study}
In MolKD, we adopt the commonly-used TAG~\cite{du2017topology} as the GNN backbone models and contrastive representation distillation~\cite{tian2019contrastive} as the knowledge distillation method. In order to show the extensibility of MolKD, we compare MolKD with other student GNN models~(GCN~\cite{kipf2016semi} and GIN~\cite{xu2018powerful}) and knowledge distillation methods(KD~\cite{hinton2015distilling}, FitNet~\cite{romero2014fitnets}, and OFD~\cite{heo2019comprehensive}).
For a fair comparison, we fix the pre-trained teacher GNN model, which achieves an AUC-ROC of $0.818$. In Table~\ref{table:ablation}, we find that the performance of MolKD is relatively stable across different student GNN models and knowledge distillation methods. The performance becomes saturated when the hidden dimension is larger than $512$. We assume the cause might be that the size of molecule data is much smaller than that of reaction data, thus the number of parameters in GNN backbone models tends to be small correspondingly.
\begin{table*}[]
\centering
\small
\caption{Performance on Tox21 among different student GNN models and knowledge distillation methods. The first and the second block in the \emph{Distillation} rows are the logit-based and feature-based knowledge distillation methods, respectively.}
\label{table:ablation}
\setlength{\tabcolsep}{12pt}
\scalebox{0.9}{
\begin{tabular}{ccccccc}
\toprule
                              &                    & TAG            & TAG            & TAG            & GCN            & GIN            \\ \midrule
\multicolumn{2}{c}{Hidden dim.}                    & 512            & 256            & 1024           & 512            & 512            \\
\multicolumn{2}{c}{\#Params (M)}                   & 0.84          & 0.49          & 2.24          & 0.81         & 1.34          \\ \midrule\midrule
\rotatebox{90}{Sup.}                          & \makecell[c]{Supervised\\student model} & $0.792_{\pm0.054}$  & $0.764_{\pm0.083}$ & $0.780_{\pm0.049}$ & $0.782_{\pm0.052}$ & $0.797_{\pm0.047}$ \\ \midrule
\multirow{4}{*}{\rotatebox{90}{Distillation}} & KD                 & $0.817_{\pm0.042}$ & $0.782_{\pm0.042}$ & $0.815_{\pm0.063}$ & $\underline{0.835_{\pm0.055}}$ & $0.797_{\pm0.040}$  \\ \cmidrule{2-7} 
                              & FitNet             & $\underline{0.830_{\pm0.030}}$ & $0.792_{\pm0.058}$ & $0.794_{\pm0.036}$ & $0.815_{\pm0.064}$ & $0.801_{\pm0.050}$ \\
                              & OFD                & $0.813_{\pm0.023}$ & $\underline{0.793_{\pm0.037}}$  & $\underline{0.826_{\pm0.026}}$  & $0.811_{\pm0.024}$ & $\underline{0.824_{\pm0.029}}$ \\
                              & MolKD           & $\bm{0.841_{\pm0.032}}$ & $\bm{0.802_{\pm0.023}}$ & $\bm{0.848_{\pm0.049}}$  & $\bm{0.837_{\pm0.038}}$ & $\bm{0.831_{\pm0.033}}$ \\ \bottomrule
\end{tabular}
}
\end{table*}

\subsection{Full Case Study on USPTO-500-MT}
\begin{figure*}[]
    \centering
    \includegraphics[width=0.62\linewidth]{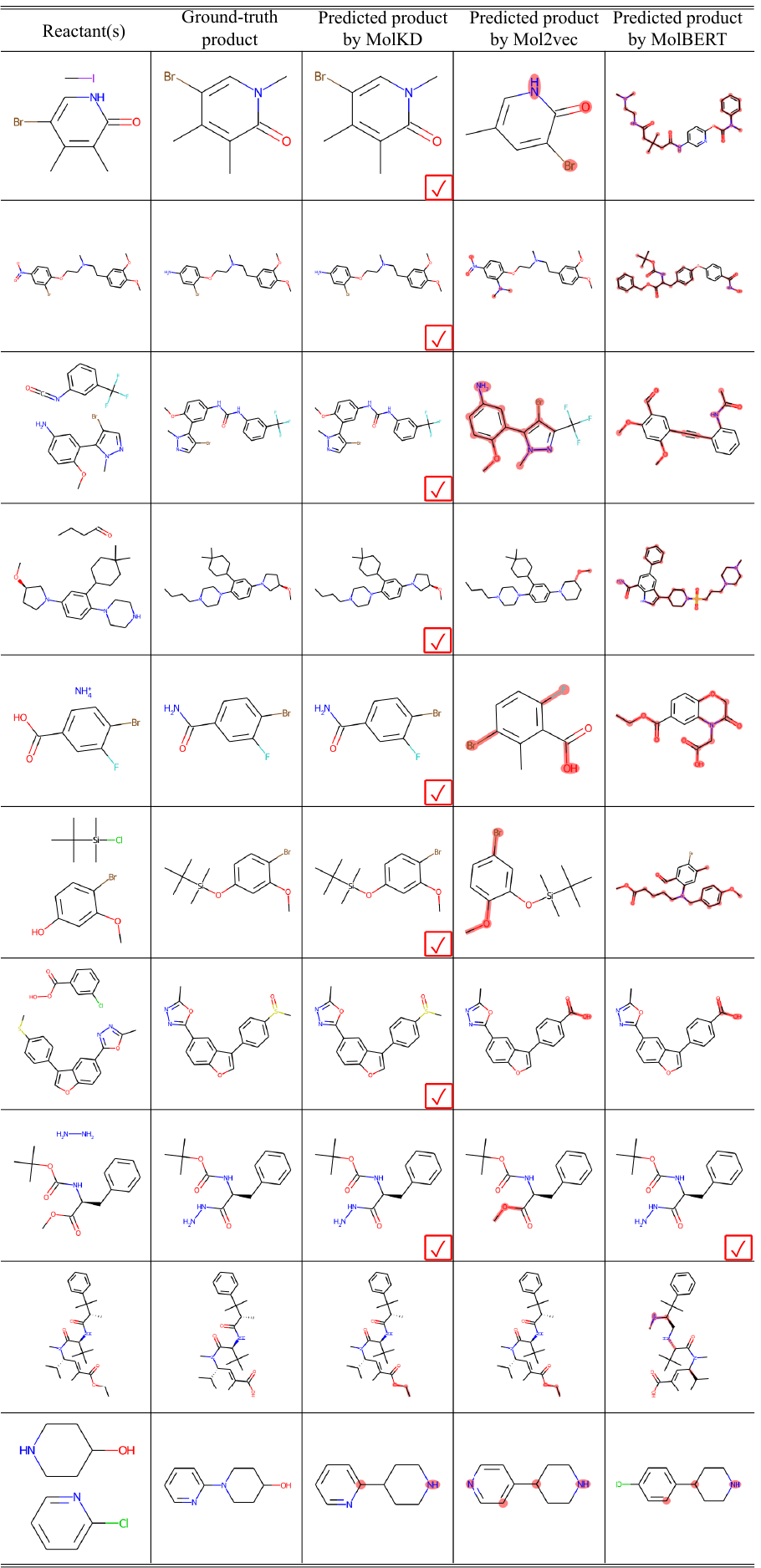}
    \caption{Case study on the USPTO-500-MT dataset. Atoms and bonds that do not match the ground-truth molecule are highlighted in red.}
    \label{fig:chemical-query-all}
\end{figure*}

We select representative reactions in the test data of USPTO-500-MT as a case study in Fig.~\ref{fig:chemical-query-all}. 
We obtain the predicted product by calculating the closest molecule with respect to the sum of molecular representations of reactants learned by the corresponding algorithm.  
We find that the predicted product by MolKD is the same as the ground-truth in most cases~(8 out of 10 cases), while Mol2vec and MolBERT fail to predict the exact product of each reaction. These results highlight the effectiveness of yield-guided chemical reaction pre-training models in MolKD. In the last two cases of Fig.~\ref{fig:chemical-query-all}, we can find that these two reactions are indeed very hard to predict and all three methods predict wrongly on some small atomic functional groups, such as -OH, and -COOH.

\subsection{Full Interpretation Cases on PhysProp}
\begin{figure*}[]
    \centering
    \includegraphics[width=0.5\linewidth]{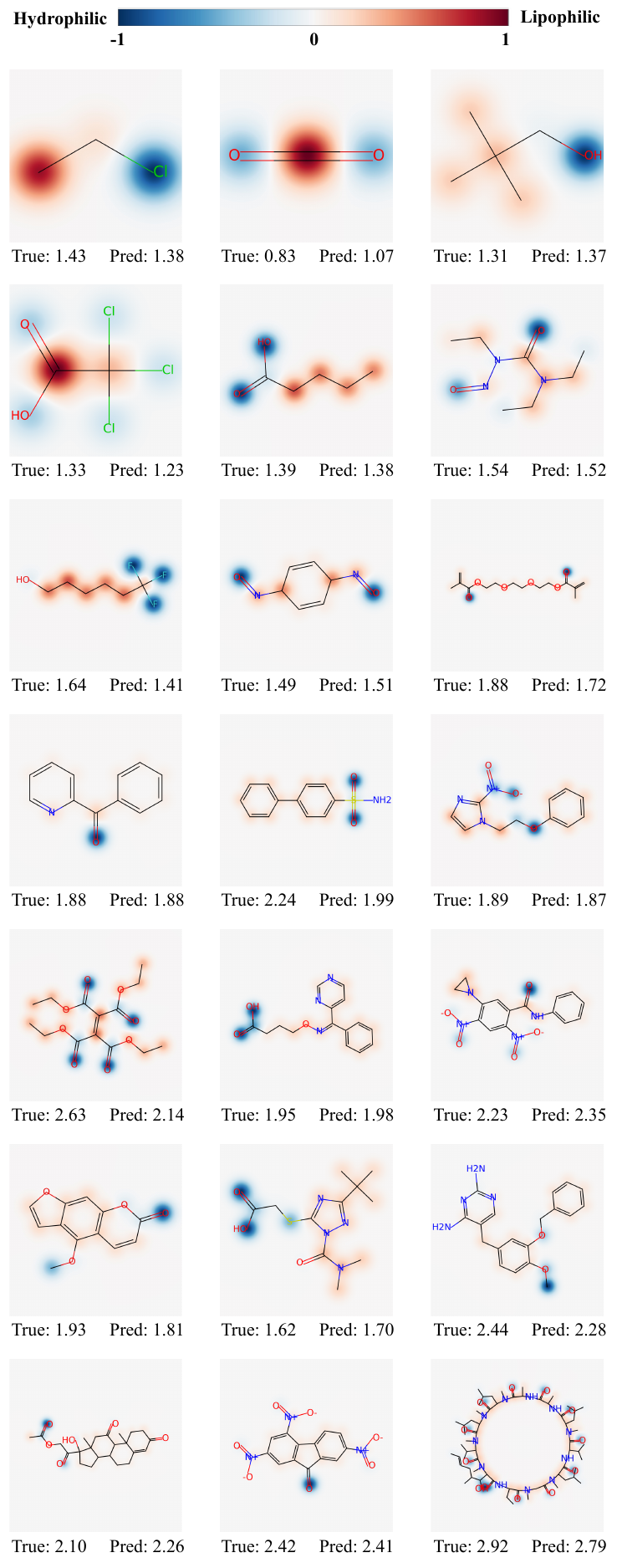}
    \caption{Case study of atom-level interpretation with true and predicted solubility labeled.}
    \label{fig:lipo-all}
\end{figure*}

Fig.~\ref{fig:lipo-all} presents a case study of solubility prediction with atom-level interpretation. These results are in with the intuition of chemists that the hydrophilic group~(\eg, -OH, and -COOH) tends to be bluer in our visualization, while the lipophilic group~(\eg, benzene ring) tends to be redder in our visualization. These observations indicate that our proposed MolKD model can distinguish essential atomic groups with clear interpretability of their solubility.

\end{document}